\documentclass[sigconf]{acmart}
\AtBeginDocument{%
  }

\usepackage{amsmath,amsfonts,bm}

\def\eqref#1{equation~\ref{#1}}

\def\1{\bm{1}}

\def\rx{{\textnormal{x}}}
\def\ry{{\textnormal{y}}}

\def\va{{\bm{a}}}
\def\vb{{\bm{b}}}

\def\vx{{\bm{x}}}
\def\vy{{\bm{y}}}

\DeclareMathAlphabet{\mathsfit}{\encodingdefault}{\sfdefault}{m}{sl}
\SetMathAlphabet{\mathsfit}{bold}{\encodingdefault}{\sfdefault}{bx}{n}

\def\gS{{\mathcal{S}}}

\def\gX{{\mathcal{X}}}
\def\gY{{\mathcal{Y}}}

\newcommand{\R}{\mathbb{R}}

\usepackage{url}
\usepackage{booktabs}
\usepackage{graphicx}
\usepackage{algorithm}
\usepackage{algpseudocode}
\usepackage{amsmath}
\usepackage{amsfonts}
\usepackage{multirow}
\usepackage{wrapfig}
\usepackage{caption}
\usepackage{subcaption}
\usepackage{pifont}
\usepackage{cleveref}
\usepackage[textsize=tiny, color=green!30]{todonotes}
\def\gray#1{\noindent\textcolor{gray}{#1}}%
\def\lgray#1{\noindent\textcolor{lightgray}{#1}}%

\def\cmark{{\color{green}\ding{51}}}%
\def\xmark{{\color{red}\ding{55}}}%

\def\method{TaRL}%
\def\metamethod{mTaRL}%

\copyrightyear{2026}
\acmYear{2026}
\setcopyright{cc}
\setcctype{by}
\acmConference[WWW '26]{Proceedings of the ACM Web Conference 2026}{April 13--17, 2026}{Dubai, United Arab Emirates}
\acmBooktitle{Proceedings of the ACM Web Conference 2026 (WWW '26), April 13--17, 2026, Dubai, United Arab Emirates}
\acmPrice{}
\acmDOI{10.1145/3774904.3792477}
\acmISBN{979-8-4007-2307-0/2026/04}

\usepackage{enumitem}

\setlist[itemize]{leftmargin=*}

\begin{document}

\title{Language Model Representations for Efficient Few-Shot Tabular Classification}


\author{Inwon Kang}
\affiliation{%
	\institution{Rensselaer Polytechnic Institute}
	\city{Troy}
	\state{New York}
	\country{USA}}
\email{kangi@rpi.edu}

\author{Parikshit Ram}
\affiliation{%
	\institution{IBM Research}
	\city{Yorktown Heights}
	\state{New York}
	\country{USA}
}
\email{Parikshit.Ram@ibm.com}

\author{Yi Zhou}
\affiliation{%
	\institution{IBM Research}
	\city{San Jose}
	\state{California}
	\country{USA}
}
\email{Yi.Zhou@ibm.com}

\author{Horst Samulowitz}
\affiliation{%
	\institution{IBM Research}
	\city{Yorktown Heights}
	\state{New York}
	\country{USA}
}
\email{samulowitz@us.ibm.com}

\author{Oshani Seneviratne}
\affiliation{%
	\institution{Rensselaer Polytechnic Institute}
	\city{Troy}
	\state{New York}
	\country{USA}}
\email{senevo@rpi.edu}

\renewcommand{\shortauthors}{Kang et al.}

\begin{abstract}
	The Web is a rich source of structured data in the form of tables, from product catalogs and knowledge bases to scientific datasets. However, the heterogeneity of the structure and semantics of these tables makes it challenging to build a unified method that can effectively leverage the information they contain. Meanwhile, Large language models (LLMs) are becoming an increasingly integral component of web infrastructure for tasks like semantic search. This raises a crucial question: can we leverage these already-deployed LLMs to classify structured  data in web-native tables (e.g., product catalogs, knowledge base exports, scientific data portals), avoiding the need for specialized models or extensive retraining? This work investigates a lightweight paradigm, \textbf{Ta}ble \textbf{R}epresentation with \textbf{L}anguage Model~(\method{}), for few-shot tabular classification that directly utilizes semantic embeddings of individual table rows. We first show that naive application of these embeddings underperforms compared to specialized tabular models. We then demonstrate that their potentials can be unlocked with two key techniques: removing the common component from all embeddings and calibrating the softmax temperature. We show that a simple meta-learner, trained on handcrafted features, can learn to predict an appropriate temperature. This approach achieves performance comparable to state-of-the-art models in low-data regimes ($k \leq 32$) of semantically-rich tables. Our findings demonstrate the viability of reusing existing LLM infrastructure for efficient semantics-driven pathway to reuse existing LLM infrastructure for Web table understanding.
\end{abstract}



\maketitle

\section{Introduction}

The Web contains a rich mixture of structured data in the form of tables -- from product catalogs and knowledge bases to scientific datasets and business intelligence dashboards. These tables often carry rich semantic information through their column names, categorical values, and textual features. As large language models (LLMs) become increasingly embedded in web infrastructure for tasks like semantic search and content understanding, a natural question arises: \textit{can we leverage these already-deployed LLM systems to understand and classify structured data, without requiring additional specialized models or extensive retraining?}

Traditional approaches to structured data classification~\cite{chenXGBoostScalableTree2016,keLightGBMHighlyEfficient2017} have relied on specialized models like gradient boosted trees or neural architectures designed specifically for tabular data. While effective, these methods do not directly leverage the semantic understanding that LLMs have developed through pre-training on massive text corpora. Recent work has explored bridging this gap through two main approaches: using LLMs as feature extractors combined with traditional classifiers~\citep{zhangELFGymEvaluatingLarge2024,kasneciEnrichingTabularData2024}, or fine-tuning LLMs on large corpora of tabular data~\citep{hegselmannTabLLMFewshotClassification2023,gardnerLargeScaleTransfer2024}. However, the first approach still requires training a downstream model, while the second demands significant computational resources (days of GPU time) and large training datasets—resources that may not be available in many practical web applications.

Moreover, existing methods that serialize entire tables for LLM processing incur prohibitive computational cost at inference time. For instance, serializing a few-shot task with just 3 columns and 2 examples can easily exceed 100 tokens, and the quadratic attention complexity of the transformer architecture makes this approachwimpractical for real-world web-scale applications. This raises a key question: \textit{can we leverage the semantic understanding already present in off-the-shelf LLM embeddings for structured data classification, without the overhead of specialized training or token-intensive serialization?}

We investigate this question through the lens of \textit{lightweight semantic understanding} -- organizations with already-deployed LLMs can reuse these systems for structured data tasks without additional infrastructure. This scenario is increasingly common: companies deploy LLMs for customer support, content generation, or semantic search, and subsequently encounter classification tasks on structured data (e.g., categorizing user-submitted forms, triaging support tickets with structured fields, or classifying product metadata). Rather than deploying a separate specialized model, can the existing LLM infrastructure be directly leveraged?
\begin{figure}
	\footnotesize
	\centering
	\includegraphics[width=0.8\linewidth]{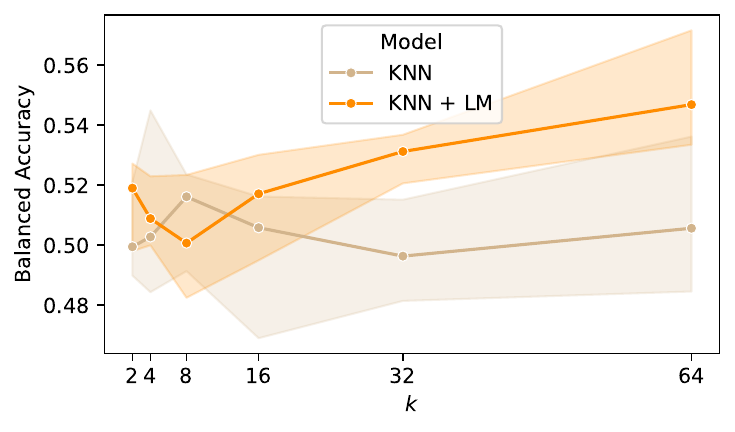}
	\captionof{figure}{Comparison between KNN on raw data and on Llama 3.1 8B's embeddings for $k$-shot classification on CARTE-CLF datasets.
		The results show that substituting raw features with LLM-derived embeddings alone yields a clear performance gain.
	}
	\label{fig:knn_vs_embedding}
\end{figure}%

Our initial experiments reveal that LLM embeddings of individual table rows do contain useful predictive signal compared to raw feature values, as shown in Figure~\ref{fig:knn_vs_embedding}. However, a naive application of these embeddings performs poorly. In this work, we identify two critical factors that prevent off-the-shelf embeddings from reaching their potential: (1) the \textit{geometry} of the embedding space, where embeddings exhibit anisotropy -- i.e. not uniformly distributed in the latent space -- and cluster in narrow regions, and (2) the need for \textit{task-specific adaptation} to account for the varying semantic structure across different classification problems.

To address these challenges, we propose \textbf{Ta}ble \textbf{R}epresentation with \textbf{L}anguage Model~(\method{}), a minimalist framework that makes LLM embeddings effective for few-shot tabular classification without any gradient-based training. Our approach requires only two lightweight adaptations: (1) a geometric correction (common component removal) to address embedding anisotropy, and (2) calibration of a single temperature parameter to adapt the classifier's behavior to each task. This second adaptation is achieved through a simple meta-learned predictor trained on a small set of auxiliary datasets, requiring less than an hour of computation compared to days required by specialized tabular foundation models.

Our method is particularly effective for semantically-rich structured data—tables with meaningful column names, categorical values, and textual features common in web applications. On such datasets, \method{} achieves competitive performance with state-of-the-art specialized models in low-data regimes ($k \leq 32$), while providing up to 1000$\times$ speedup over autoregressive LLM inference. This makes it especially valuable for web-based applications where LLMs are already deployed and tables contain rich semantic information that can be leveraged for classification tasks.

Our work demonstrates that lightweight semantic understanding of structured data can be achieved by reusing existing LLM infrastructure, without requiring specialized architectures or extensive retraining. The key lies not in building more complex models, but in properly adapting the semantic representations that LLMs have already learned.~\Cref{tab:related_works} shows a comparison of our approach with existing tabular foundation models, highlighting our method's unique advantages in terms of model agnosticism and minimal overhead.

Our contributions are as follows:
\begin{itemize}
	\item We introduce \method{}, a lightweight framework that reuses off-the-shelf LLM semantic embeddings of serialized table rows for few-shot classification on web datasets, requiring no gradient-based training and minimal computational overhead.
	\item We identify two key semantic adaptations: Common Component Removal (CCR) to mitigate embedding anisotropy and per-task temperature calibration to enable dynamic attention over context examples, and show they unlock much of the performance latent in LLM embeddings.
	\item We establish an oracle upper bound for perfect per-task calibration and approximate it with a simple meta-learned predictor that trains in matters of minutes, enabling practical deployment.
	\item We demonstrate competitive accuracy and dramatic runtime gains (up to $1000\times$ under caching) on semantically-rich datasets of CARTE-CLF benchmarks in the low-data regime ($k\leq 32$), highlighting where table semantics wins.
\end{itemize}

\begin{table*}
	\centering
	{\small
		\begin{tabular}[c]{lccccc}
			\toprule
			                                                                      & Semantic   & No Downstream & No Task & Model    & Table-Tuning                   \\
			                                                                      & Embeddings & Training      & Prompt* & Agnostic & Overhead$^\dagger$             \\ \midrule
			\textbf{TabLLM}~\cite{hegselmannTabLLMFewshotClassification2023}      & \cmark     & \xmark        & \xmark  & \xmark   & \gray{Not Reported}$^\ddagger$ \\
			\textbf{TabPFN}~\cite{hollmannAccuratePredictionsSmall2025}           & \xmark     & \cmark        & \cmark  & \xmark   & 2 weeks                        \\
			\textbf{ConTextTab}~\cite{spinaciConTextTabSemanticsAwareTabular2025} & \cmark     & \cmark        & \cmark  & \xmark   & 4-10 days                      \\
			\textbf{TabuLa-8B}~\cite{gardnerLargeScaleTransfer2024}               & \cmark     & \cmark        & \xmark  & \xmark   & 6 days                         \\ \midrule
			\textbf{\method{}} (ours)                                             & \cmark     & \cmark        & \cmark  & \cmark   & $<$ 1 hour                     \\
			\bottomrule
		\end{tabular}
	}
	\caption{
		Comparison of our proposed approach (\method{}) against existing tabular
		foundation models. TabPFN and ContextTab are specialized architectures
		designed specifically for tabular prediction task, while TabuLa-8B is an
		LLM fine-tuned specifically for few-shot tabular classification. In
		contrast, \method{} uses off-the-shelf LLM embeddings without any
		task-specific training or prompting.\\
		\scriptsize{ *: Some methods require adding
			additional task-specific context to the input (e.g. column names, feature
			descriptions, etc.) in addition to the context examples to achieve good
			performance.\\ $^\dagger$: The cost of pre-training for \textit{tabular}
			tasks -- only the table-training overhead is considered for TabuLa-8B.\\
			$^\ddagger$: The authors report fine-tuning language models, but do not specify the exact
			runtime.}
	}
	\label{tab:related_works}
\end{table*}

\begin{figure}
	\centering
	\includegraphics[width=\linewidth]{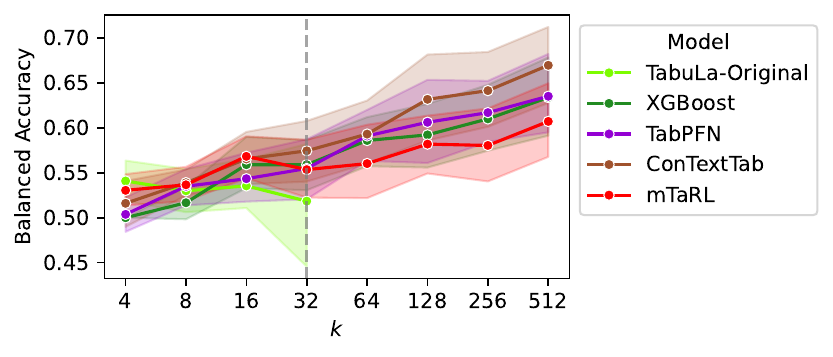}
	\vspace{-0.8cm}
	\captionof{figure}{
    Comparison of \metamethod{} against SotA baselines on CARTE-CLF benchmark (11 datasets). \metamethod{} shows competitive performance against SotA models in lower $k$ settings. Gray line indicates the context limit for TabuLa, and is able to scale much further than the 32-shot limit imposed for TabuLa. 
	}
	\label{fig:main_comparison}
\end{figure}

\section{Related Work}

\subsection{Table Understanding and Semantic Embeddings}

It has been shown that learning effective representations for tabular data can significantly improve performance on downstream tasks~\citep{dengTURLTableUnderstanding2020,guoEntityEmbeddingsCategorical2016,badaroTransformersTabularData2023,zengLLMEmbeddingsImprove2024}. Previous works have explored various methods of learning such embeddings, ranging from task-specific learned embeddings~\cite{zhangTable2VecNeuralWord2019,wu2024deep,kang2025on} to pre-training strategies~\cite{iidaTABBIEPretrainedRepresentations2021,yangUniTabEUniversalPretraining2023,spinaciPORTALScalableTabular2024}. The resulting embeddings are then often used as input features for the downstream task, such as retrieval~\cite{berenguerWordEmbeddingsRetrieving2024,cappuzzoCreatingEmbeddingsHeterogeneous2020,zhengDenseRepresentationLearning2023} or classification~\cite{huangTabTransformerTabularData2020,koloskiLLMEmbeddingsDeep2025}.

\subsection{Classical and Specialized Tabular Models}

For many years, Gradient Boosted Decision Trees (GBDTs), such as XGBoost~\citep{chenXGBoostScalableTree2016} and LightGBM~\cite{keLightGBMHighlyEfficient2017}, have been the state-of-the-art for tabular data, particularly in industrial applications and data science competitions. While deep learning models like TabNet~\citep{arikTabNetAttentiveInterpretable2021} and FT-Transformer~\citep{gorishniyRevisitingDeepLearning2021} have demonstrated competitive performance, GBDTs remain a strong and widely used baseline.

More recently, the field has seen the emergence of specialized foundation models for tabular data. TabPFN~\citep{hollmannAccuratePredictionsSmall2025} introduced a prior-fitted network that uses a transformer to approximate Bayesian inference, achieving strong performance on small to medium sized tables without task-specific fine-tuning. ConTextTab~\citep{spinaciConTextTabSemanticsAwareTabular2025} extended this approach by incorporating semantic information from column names, further improving performance. These models, while powerful, are specialized architectures designed specifically for tabular data. In contrast, our work investigates whether it is possible to achieve strong performance without resorting to specialized models, by instead adapting the general-purpose representations from off-the-shelf LLMs.

\subsection{Large Language Models for Tabular Data}

The use of LLMs for tabular data has been explored through several paradigms. One common approach is to use LLMs to generate feature embeddings for tabular data, which are then fed into a separate downstream classifier. For instance, \citet{zhangELFGymEvaluatingLarge2024} and \citet{kasneciEnrichingTabularData2024} have shown that adding LLM-generated embeddings as additional features for models like XGBoost can improve performance.~\citet{hanLargeLanguageModels2024} shows that LLM engineered features are especially powerful in few-shot settings. However, this still requires training a separate, often complex, downstream model or employs sophisticated (and often token intensive) prompting techniques. Our work diverges from this by proposing an approach that directly uses the LLM embeddings for classification without any gradient-based training.

Another line of research focuses on adapting LLMs to tabular data through fine-tuning or in-context learning. TabLLM~\citep{hegselmannTabLLMFewshotClassification2023} demonstrated that LLMs like GPT-3 can perform in-context learning for tabular tasks with carefully designed prompts. LIFT~\citep{dinhLIFTLanguageInterfacedFineTuning2022a} improved few-shot performance by fine-tuning on datasets retrieved based on a new task description. More recent work, such as~\citet{yangUnleashingPotentialLarge2025}, Table-GPT~\citep{liTableGPTTableFinetuned2024} and TabuLa-8B~\citep{gardnerLargeScaleTransfer2024}, has focused on large-scale fine-tuning of LLMs on millions of tables to instill strong tabular reasoning capabilities. While these methods have achieved impressive results, they are computationally intensive and require significant training resources. On the other hand, model-reprogramming~\cite{chen2022model}, which repurposes pre-trained models for new tasks without modifying the model weights, can become a more efficient alternative by changing how the input data is presented to the model. This is typically done by learning a mapping from the new task's input space to the model's original input space. While this approach has been explored in various domains, its application to tabular data remains limited. Our work aims to fill this gap by demonstrating that with minimal adaptations, off-the-shelf LLM embeddings can be effectively used for tabular classification tasks, providing a lightweight alternative to the more resource-intensive methods mentioned above.

\subsection{Geometry of Language Model Representations}

Language model embeddings have been widely studied, particularly in the context of natural language processing tasks.~\citet{behnamghaderLLM2VecLargeLanguage2024} show that decoder-only LLMs can be used to generate powerful text embeddings. However, it is well-established that embeddings from transformer-based language models exhibit anisotropy, meaning they occupy a narrow cone in the vector space~\citep{ethayarajhHowContextualAre2019}. This can degrade the performance of downstream tasks that rely on distance metrics like cosine similarity. A common technique to mitigate this is to post-process the embeddings, for example, by subtracting the common component~\citep{aroraSimpleToughtoBeatBaseline2017,muAllbuttheTopSimpleEffective2018}. Our work builds on this understanding, but we go a step further by demonstrating that a simple, context-specific geometric correction is a critical component for unlocking the information contained in LLM embeddings for tabular data. While more sophisticated methods for correcting representation geometry exist, we find that a simple and targeted approach is sufficient to achieve a significant performance improvement.

\section{Methodology}

\begin{table}
	\centering
	\captionof{table}{Comparison of averaged runtime (in seconds) per $k$-shot task using TabuLa-8B with \method{} and original auto-regressive way on CARTE-CLF benchmark.}
	\label{tab:main_runtime_comparison}
	\begin{tabular}[c]{llll}
		\toprule
		$k$ & \textbf{TabuLa} & \textbf{\method{}}(Ours)            \\ \midrule
		2   & 36867.90        & 29.93 {\color{blue}($\times 1232$)} \\
		4   & 37518.86        & 42.19 {\color{blue}($\times 889$)}  \\
		8   & 34785.27        & 67.19 {\color{blue}($\times 518$)}  \\
		16  & 30369.63        & 118.06 {\color{blue}($\times 257$)} \\
		32  & 21455.61        & 222.76 {\color{blue}($\times 96$)}  \\
		\bottomrule
	\end{tabular}
\end{table}%

We hypothesize that two key factors can significantly unlock the potential of these embeddings: (1)~a geometric correction to account for the anisotropy of the embedding space, and (2)~task-specific adaptation of the classifier to account for differences in data distributions across tasks.
Our method, \textbf{\method{}}, avoids any gradient-based training to isolate the predictive power of the LLM embeddings themselves, as well any task-specific tuning. It consists of four main components: a serialization function to convert tabular rows into text, a kernel-based classifier that operates on the LLM embeddings, a geometric correction to the embedding space, and softmax-temperature scaling calibration to adapt to different tasks. We describe each component in detail below. We will use $\rx$ to denote a row with $d$ features, and $\vx \in \mathbb{R}^{d_e}$ to denote the corresponding LLM embedding with $d_e$ components.

\subsection{From Tabular Rows to LLM Embeddings}

To make tabular data accessible to a language model, we first serialize each row into a text sequence. We define a serialization function, $s = \text{serialize}(\rx)$, which converts a row $\rx \in \mathbb{R}^{d_\text{num}} \times \mathbb{C}^{d_\text{cat}}$ into a human-readable string. This is done by concatenating the column names and their corresponding values. For example, a row with columns \texttt{Feature A} and \texttt{Feature B} and values 1.2 and 3.4 would be serialized as: \texttt{Feature A: 1.2, Feature B: 3.4}. Similarly, categorical features are encoded using their string representation (if available), and date/time features are converted into a standard format (\texttt{YY-MM-DD:HH}). Each row $\rx$ is then embedded into a vector $\vx \in \R^{d_e}$ using a pretrained, off-the-shelf LLM. It is worth noting that each row is embedded \textit{separately} in ~\method{}. This is important to note because the cost of modern transformer-based embedding methods scales quadratically with the length of the input sequence~\citep{attention_all_you_need}. Thus, unlike previous methods where the entire context and the task prompt was serialized into a single input, the cost of embedding for~\method{} will only scale quadratically with respect to the \textit{single longest context example}.

\subsection{\method{}: Few-Shot Classification via Kernel Regression}
\label{sec:kernel_regression}

We frame the few-shot classification problem as a kernel regression task. Given a support set of $k$ labeled examples, $D_\text{supp} = \{(\rx_i, \ry_i)\}_{i=1}^k$ and set of $m$ possible label classes $\gY$, where $\rx_i$ is a tabular data point and $\ry_i \in \gY$ is its class label, our goal is to predict the label $\hat{\ry}_q$ for a new query point $\rx_q$. We choose a kernel-based approach because it is non-parametric, makes few assumptions about the underlying data distribution, and is well-suited for the low-data regime of few-shot learning.

First, we compute the embeddings for the query and all support examples, yielding $\vx_q$ and $\{\vx_i\}_{i=1}^k$. Next, we compute the similarity $\text{sim}(\vx_q, \vx_i)$ between the query embedding $\vx_q$ and each support embedding $\vx_i$ using a similarity metric. We use the negative inverse cosine similarity, defined as: $\text{sim}(\va, \vb) = -\cos^{-1}(\va, \vb)$, i.e. the angle between the two vectors. We opt to use this metric because it is bounded between $[-\pi, 0]$, which helps stabilize the softmax computation in the next step. These similarities are then normalized via a softmax function to produce a set of weights, $w_i$:

\begin{equation}
	w_i = \frac{\exp(\text{sim}(\vx_q, \vx_i))}{\sum_{j=1}^k \exp(\text{sim}(\vx_q, \vx_j))}
	\label{eq:weights}
\end{equation}

Then, the query's label embedding is computed as a weighted sum of the support set labels.

\begin{equation}
	\bar{\vy}_q = \sum_{i=1}^k w_i \vy_i
\end{equation}

Finally, the the predicted class $\hat{y}_q$ is determined by comparing $\bar{\vy}_q$ against the possible label class embeddings $\{\vy_c\}_{c \in \gY}$, where $\vy_c$ is the embedding of class $c$. The predicted class is the one with the highest similarity to $\bar{\vy}_q$. It is worth noting here that it is also possible to use a one-hot vector of the labels in place of the embedding $\vy_i$. We later examine the effect of both approaches.

\begin{equation}
	\hat{y}_q = \arg\max_{c \in \gY} \text{sim}(\bar{\vy}_q, \vy_c)
\end{equation}

This simple classifier allows us to directly probe the quality of the LLM embeddings without introducing any trainable parameters.

\subsection{Geometric Correction with Common Component Removal (CCR)}

As noted earlier, the geometry of the LLM embedding space can be a significant obstacle to good performance. Specifically, many language-embedding techniques are known to suffer from \textit{anisotropy}, where the embeddings are not uniformly distributed in space, but rather clustered around a few dominant directions~\citep{ethayarajhHowContextualAre2019}. This can potentially lead to poor discrimination between different classes, as the embeddings may not be well-separated. In the context of our kernel regression classifier, this anisotropy can cause the similarity scores to be dominated by these common directions, leading to suboptimal weight distributions and ultimately poor classification performance. To address this, we apply a simple but effective geometric correction called Common Component Removal (CCR). This technique, adapted from work on sentence embeddings~\citep{aroraSimpleToughtoBeatBaseline2017,muAllbuttheTopSimpleEffective2018}, is designed to improve the isotropy of the embedding space by removing the average embedding vector (the common component) from all embeddings. We find that applying this technique (lines 6 and 7 of Algorithm~\ref{alg:tarplm_base_compact}) significantly improves performance across all datasets and few-shot settings.

\subsection{Calibrating the Softmax Temperature}

So far, the kernel classifier maps similarities directly to weights via a softmax function (Eq.~\ref{eq:weights}). However, the scale of the similarities can vary significantly across different tasks and datasets -- e.g., in some tasks, the similarities may be tightly clustered, while in others they may be more spread out. To account for this variability, we introduce a tunable parameter $\gamma$ to control the sharpness of the softmax distribution.

\begin{equation}
	\label{eq:gamma_weights}
	w_i = \frac{\exp(\gamma \cdot \text{sim}(\vx_q, \vx_i))}{\sum_{j=1}^k \exp(\gamma \cdot \text{sim}(\vx_q, \vx_j))}
\end{equation}

This parameter $\gamma$ can be interpreted as an inverse of the temperature parameter that adjusts the confidence of the weight distribution. Thus, a higher value of $\gamma$ leads to a more \textit{peaked} distribution, essentially mimicking a nearest-neighbor classifier, while a lower value leads to a more \textit{uniform} distribution, effectively averaging over more support examples. In the following section, we show the careful tuning of $\gamma$ allows the classifier to leverage the LM embeddings more effectively. Algorithm~\ref{alg:tarplm_base_compact} summarizes the full \textbf{\method{}} prediction procedure, which combines all the components described above.

\paragraph{A heuristic for setting $\gamma$.} In practice, we find that the optimal value of $\gamma$ varies significantly across datasets and few-shot settings. This means that we need a way to select $\gamma$ by just looking at the support set. One such heuristic is to set $\gamma$ based on the \textit{how similar the most similar support example is} -- i.e., the maximum similarity between the query and any support example.

\begin{equation}
	\hat{\gamma} = \frac{1}{\max\left(-\text{sim}(\vx_q, \vx_i)\right)}
	\label{eq:gamma_heuristic}
\end{equation}

If the similarity of the closest example is very high (i.e., the closest example is a near-duplicate of the query), $\hat{\gamma}$ will approach infinity, meaning Eq.~\ref{eq:gamma_weights} effectively puts all the weight on the nearest-neighbor, and thus the final classification matches that of a one-nearest-neighbor classifier. On the other hand, as the similarity between the query and the closest example decreases, $\hat{\gamma}$ will cause the weight to be more evenly distributed across every sample in the context.

\begin{algorithm}[H]
	{\small
		\caption{\textbf{\method{}} Few-Shot Prediction}
		\label{alg:tarplm_base_compact}
		\begin{algorithmic}[1]
			\State \textbf{Input:} Support set $\gS = \{(\rx_i, y_i)\}_{i=1}^k$, Query row $\rx_q$, Temperature $\gamma$, Label classes $\gY$
			\State $\vx_i \leftarrow \textsc{LM}(\textsc{Serialize}(\rx_i))$; $\vy_i \leftarrow \textsc{LM}(y_i)$ for $i \in \{1..k\}$
			\State $\vx_q \leftarrow \textsc{LM}(\textsc{Serialize}(\rx_q))$
      \State $\vy_c \leftarrow \textsc{LM}(c)$ for $c \in \gY$
			\State $\bar{\vx} \leftarrow \frac{1}{k+1} \left((\sum_{i=1}^k \vx_i) + \vx_q\right)$
			\State $\vx'_i \leftarrow \vx_i - \bar{\vx}$ for $i \in \{1..k\}$ \quad {\color{gray}\footnotesize\# Apply CCR to support}
			\State $\vx'_q \leftarrow \vx_q - \bar{\vx}$ \quad {\color{gray}\footnotesize\# Apply CCR to query}
			\State Compute weights $\{w_i\}_{i=1}^k$ for $(\vx'_q, \{\vx'_i\}_{i=1}^k)$ using Eq.~\ref{eq:gamma_weights}.
			\State $\bar{\vy}_q \leftarrow \sum_{i=1}^k w_i \vy_{y_i}$, where $\vy_{y_i}$ is the embedding for class label $y_i$.
			\State $\hat{y}_q \leftarrow \arg\max_{c \in \gY} \text{sim}(\bar{\vy}_q, \vy_c)$
			\State \textbf{Return:} Predicted label $\hat{y}_q$
		\end{algorithmic}
	}
\end{algorithm}

\subsection{\metamethod{}: Learning to calibrate $\gamma$}

While the heuristic-based method provides a simple way to set $\gamma$, it is not guaranteed to be optimal. To further improve performance, we propose a prototype meta-learning approach, \metamethod{}, that learns to predict the optimal $\gamma$ from the properties of the support set itself. We first define an \textit{oracle} classifier: a classifier that uses the optimal $\gamma$ for each task, where the optimal $\gamma$ is chosen by minimizing the cross-entropy loss on the query's true label. This oracle classifier serves as an upper bound on the performance of our method, as it represents the best possible performance we could achieve if we knew the optimal $\gamma$ for each task. We then approximate this oracle by training a simple regression model on a set of general, dataset-agnostic meta-features. These features are not engineered for any specific dataset but rather capture only the task-relevant context-query statistics of the embedding space, such as mean similarity, variance, and class imbalance (see \Cref{sec:apdx:meta_learnre_details} for details). This lightweight, support-set-level feature extraction allows for rapid, gradient-free adaptation to new tasks and stands in stark contrast to computationally intensive alternatives like LLM fine-tuning. The model is trained on datasets with known optimal $\gamma$ values, determined via cross-validation.

Given a new few-shot task, we compute the same summary statistics from its support set and use the trained regression model to predict the optimal $\gamma$. This approach allows us to adaptively select $\gamma$ for each new task without any gradient-based training on the task itself. We find that this meta-learning approach consistently outperforms the heuristic method, demonstrating the effectiveness of learning to adapt the classifier to different tasks.

Algorithms~\ref{alg:meta_datagen_compact} and~\ref{alg:meta_inference_compact} summarize the meta-training and meta-inference procedures, respectively. The meta-training procedure constructs a dataset of support set features and optimal $\gamma$ values from multiple datasets and few-shot settings. The optimal $\gamma$ is chosen by picking the $\gamma$ that maximizes the probability of the correct prediction on a held-out query point. This is done for multiple tasks sampled from each dataset to create a diverse meta-training set. The features are extracted \textit{only} from the support set using a feature engineering function $f$. More detail on $f$ can be found in~\Cref{sec:apdx:meta_learnre_details}.
The meta-inference procedure trains a regression model on this meta-dataset and uses it to predict $\bar{\gamma}$ for new tasks. This allows us to adaptively select $\gamma$ for each new task without any gradient-based training on the task itself.


\section{Experiments}

\subsection{Datasets}

We evaluate our method on a suite of datasets selected from the CARTE benchmark~\citep{kimCARTEPretrainingTransfer2024} and TabArena benchmark~\citep{ericksonTabArenaLivingBenchmark2025}. 

The CARTE benchmark, consisting of 11 binary classification and 40 regression tasks, is curated with a focus on datasets that have \textit{low numerical features} and \textit{high-cardinality textual features}, making it well suited for evaluating tabular models that can handle semantic content. Additionally, the datasets in this benchmark have meaningful column names, which are important for LLMs to effectively utilize the tabular data. 

TabArena is a more recent benchmark that includes 32 binary, 6 multiclass classification tasks\footnote{We discard 4 classification datasets due to issues with large memory usage.} and 13 regression tasks, with a focus on datasets that best represent the most \textit{general} tabular classification task. Unlike CARTE, datasets in TabArena may not have meaningful column names and can include a higher proportion of numerical features. 

A detailed report on the dataset characteristics can be found in~\Cref{tab:dataset_statistics} in~\Cref{sec:apdx:meta_learnre_details}. In total, we evaluate our approach using 45 classification datasets and 53 regression datasets. Our main results focus on the binary classification datasets of each benchmark as~\method{} is a classification mechanism. Additionally, we also report the results on the multiclass classification datasets from TabArena and binned versions of the regression datasets in~\Cref{sec:apdx:detailed_results}. We use all 98 of the datasets for generating the meta-dataset for training the meta-learner.

\subsection{Sampling Procedure}

To simulate a few-shot learning scenario, we randomly sample a small number of examples from the training set to form the support set. We vary the size of the support set, $k$, from 2 to 512, using the following values: $k \in \{2, 4, 8, 16, 32, 64, 128, 256, 512\}$. To ensure a fair comparison, we maintain the class balance of the original dataset when sampling the support set. For each value of $k$, we repeat the experiment with 128 different randomly sample context-query pairs. It is worth noting that the baseline few-shot results reported in~\citet{gardnerLargeScaleTransfer2024} were obtained by optimizing the context and re-using it for every query. However, this appraoch makes the assumption that \textit{we can choose the best possible context for each task}, which is not realistic in practice. Thus, in order to simulate a realistic low-data scenario, we opt to re-sample the support set for each query, which makes the task more challenging and realistic. All the results reported in this work are based on this sampling procedure for a fair comparison.

\subsection{Models and Baselines}

We compare our method against three categories of baselines:

\begin{itemize}
	\item \textbf{Traditional Tabular Learners:} We include standard, widely-used models for tabular data: XGBoost~\citep{chenXGBoostScalableTree2016} and k-Nearest Neighbors (KNN). These models are trained on the support set and evaluated on the corresponding query row. Since the context size is quite small, we do not apply any hyperparameter optimization, aside from setting the number of neighbors in KNN to $\log_2(k)$.
	\item \textbf{Specialized Tabular Foundation Models:} We compare against state-of-the-art foundation models designed specifically for tabular data: TabPFN~\citep{hollmannAccuratePredictionsSmall2025}, and ConTextTab~\citep{spinaciConTextTabSemanticsAwareTabular2025}, a semantics-enhanced version of the PFN backbone. These models are used off-the-shelf without any fine-tuning.
	\item \textbf{LLM-based Models:} We compare against other methods that use LLMs for tabular data. This includes TabuLa-8B~\citep{gardnerLargeScaleTransfer2024}, Llama 3.1 8B variants~\citep{grattafiori2024llama3herdmodels}, Granite 3 8B~\citep{granite3} and Qwen 3 8B~\citep{yang2025qwen3technicalreport}. In addition, we also test a smaller encoder-only model from Sentence Transformers miniLM-L6-v2~\citep{reimersSentenceBERTSentenceEmbeddings2019}.
\end{itemize}

For picking the \textit{optimal} performance and constructing the meta-dataset, we choose values of $\gamma$ from $2^{[-10, 15]}$.
We also consider using the LLM in an auto-regressive manner as used by previous works~\citep{gardnerLargeScaleTransfer2024,hegselmannTabLLMFewshotClassification2023} to directly perform few-shot classification via in-context prompting. However, as~\Cref{tab:main_runtime_comparison} shows, the prohitive resource usage of this approach made it difficult to run on all datasets. Thus, for TabuLa-8B, we opt to spend our available computing budget on CARTE-CLF for $k \leq 32$, whose semantic content should be well-suited for LLMs and would provide a fair baseline\footnote{Gathering all the results for TabuLa-8B took approximately 12 days on our setup reported in~\Cref{sec:apdx:implementation_details}.}. In addition, we also use the embeddings of TabuLa-8B for~\method{} unless otherwise specified.

\paragraph{Evaluation Metric.} In order to account for true performance with respect to label imbalance, we use balanced accuracy as our primary evaluation metric. Balanced accuracy is defined as the average of recall obtained on each class. This metric is particularly important in few-shot settings where class imbalance can significantly skew results.

\section{Results and Analysis}

\begin{table*}
	\centering
	\caption{
		Detailed comparison of average balanced accuracy score per benchmark. Best performance in each benchmark/$k$ combination is marked with \textbf{bold} and the second best with \underline{underline}. CB: CARTE-Binary, TB: TabArena-Binary. The values inside parentheses indicate the number of datasets per benchmark.
	}
	\label{tab:detailed-comparison}
	{\small
		\begin{tabular}{ll p{0.8cm}p{0.8cm} *{8}{r}}
			\toprule
			\multirow{2}{*}{Benchmark} & \multirow{2}{*}{Model} & \multicolumn{2}{c}{Rank} & \multicolumn{8}{c}{$k$ (number of shots)}                                                                                                                                                                         \\
			                           &                        & $k \leq 32$              & $k > 32$                                  & 4                  & 8                  & 16                 & 32                 & 64                 & 128                & 256                & 512                \\
			\cmidrule(lr){1-2} \cmidrule(lr){3-4} \cmidrule(lr){5-12}

			\multirow{8}{*}{\shortstack{CARTE                                                                                                                                                                                                                                                                  \\
					Binary (11)}}
			                           & KNN                    & 8.0568                   & 7.7614                                    & 0.5085             & 0.5094             & 0.5026             & 0.4993             & 0.4965             & 0.5056             & 0.5179             & 0.5096             \\
			                           & KNN-Emb.               & 7.3977                   & 7.2955                                    & 0.5125             & 0.5216             & 0.5317             & 0.5517             & 0.5450             & 0.5612             & 0.5645             & 0.5879             \\
			                           & XGBoost                & 6.8295                   & 6.1250                                    & 0.5000             & 0.5167             & 0.5590             & \underline{0.5590} & 0.5860             & 0.5922             & 0.6100             & 0.6338             \\
			                           & TabPFN                 & 6.9318                   & \underline{5.6364}                        & 0.5038             & 0.5350             & 0.5434             & 0.5548             & \underline{0.5910} & \underline{0.6062} & \underline{0.6170} & \underline{0.6352} \\
			                           & ConTextTab             & \textbf{6.3295}          & \textbf{5.0795}                           & 0.5160             & \textbf{0.5390}    & \underline{0.5655} & \textbf{0.5745}    & \textbf{0.5931}    & \textbf{0.6318}    & \textbf{0.6415}    & \textbf{0.6697}    \\
			                           & TabuLa-Original        & 6.5114                   & \lgray{N/A}                               & \textbf{0.5409}    & 0.5300             & 0.5354             & 0.5185             & \lgray{N/A}        & \lgray{N/A}        & \lgray{N/A}        & \lgray{N/A}        \\
			                           & \metamethod{}          & \underline{6.4886}       & 6.6705                                    & \underline{0.5305} & \underline{0.5367} & \textbf{0.5683}    & 0.5535             & 0.5603             & 0.5819             & 0.5804             & 0.6072             \\
			                           & \gray{Oracle}          & \gray{0.5568}            & \gray{1.0000}                             & \gray{0.7742}      & \gray{0.8057}      & \gray{0.8220}      & \gray{0.8243}      & \gray{0.8260}      & \gray{0.8469}      & \gray{0.8481}      & \gray{0.8335}      \\ \midrule

			\multirow{7}{*}{\shortstack{TabArena                                                                                                                                                                                                                                                               \\
					Binary (28)}}
			                           & KNN                    & 6.3973                   & 7.4866                                    & 0.5355             & 0.5663             & 0.6160             & 0.6072             & 0.6081             & 0.6347             & 0.6393             & 0.6216             \\
			                           & KNN-Emb.               & 7.5089                   & 7.6875                                    & 0.5250             & 0.5495             & 0.5359             & 0.5573             & 0.6138             & 0.6230             & 0.6524             & 0.6349             \\
			                           & XGBoost                & 6.1071                   & 5.8036                                    & 0.5179             & 0.5770             & 0.6136             & 0.6678             & 0.6839             & 0.7003             & 0.7153             & 0.7240             \\
			                           & TabPFN                 & \textbf{4.7545}          & \textbf{4.1875}                           & \underline{0.5688} & \textbf{0.6126}    & \textbf{0.6507}    & \textbf{0.6961}    & \textbf{0.7141}    & \textbf{0.7499}    & \underline{0.7700} & \underline{0.7526} \\
			                           & ConTextTab             & \underline{5.4152}       & \underline{4.5268}                        & \textbf{0.5691}    & \underline{0.5944} & \underline{0.6352} & \underline{0.6747} & \underline{0.7008} & \underline{0.7363} & \textbf{0.7726}    & \textbf{0.7544}    \\
			                           & \metamethod{}          & 6.7411                   & 7.3438                                    & 0.5475             & 0.5712             & 0.5744             & 0.5974             & 0.6263             & 0.6311             & 0.6409             & 0.6416             \\
			                           & \gray{Oracle}          & \gray{0.5312}            & \gray{1.1071}                             & \gray{0.8038}      & \gray{0.8197}      & \gray{0.8397}      & \gray{0.8551}      & \gray{0.8776}      & \gray{0.8862}      & \gray{0.8932}      & \gray{0.8741}      \\

			\bottomrule
		\end{tabular}
	}
\end{table*}

The main results are summarized in Table~\ref{tab:detailed-comparison}, which compares our proposed meta-learning method (\metamethod{}) against several baselines across different benchmarks and varying numbers of shots ($k$). First, we see that the oracle classifier yields extremely strong performance, achieving the best or near-best results across all benchmarks and $k$ values. It is important to note that this performance is not necessarily an achievable upper bound as it relies on an oracle to select the best $\gamma$ for each instance. However, it demonstrates the potential of LLM embeddings when properly adapted.

\begin{table}
	\centering
	\caption{Comparison of average balanced accuracy score when using label embeddings (Oracle) vs. one-hot encoded labels (Oracle-V).}
	\label{tab:oracle-label-comparison}
	{
		\small
		\setlength{\tabcolsep}{2.5pt}
		\begin{tabular}{l *{8}{r}}
			\toprule
			\multirow{2}{*}{Model} & \multicolumn{8}{c}{$k$ (number of shots)}                                                                \\
			                       & 4                                         & 8      & 16     & 32     & 64     & 128    & 256    & 512    \\
			\midrule
			Oracle                 & 0.7742                                    & 0.8057 & 0.8220 & 0.8243 & 0.8260 & 0.8469 & 0.8481 & 0.8335 \\
			Oracle-V               & 0.5869                                    & 0.6360 & 0.6884 & 0.7041 & 0.7490 & 0.7703 & 0.7882 & 0.7892 \\
			\bottomrule
		\end{tabular}
	}
\end{table}%

\subsection{Predictive Signal in Raw LLM Embeddings}

Our first analysis is to examine whether the raw LLM embeddings contain any useful signal for tabular classification. While previous works have examined using LLM embeddings as \textit{additional} features to the tabular data, we are interested in whether they can be used \textit{directly} for classification without any additional features or training. To this end, we compare a simple KNN classifier using the raw tabular data against a KNN classifier using the LLM embeddings from TabuLa-8B using $log_2(k)$ as the neighbor size for each $k$-shot task.~\Cref{fig:knn_vs_embedding} shows the results on CARTE-CLF benchmark. We find that that 1). KNN on raw data performs poorly in low $k$ settings, and 2). KNN on LLM embeddings performs significantly better, demonstrating that the LLM embeddings do indeed contain useful information for tabular classification. However, the performance is still far from satisfactory, indicating that further adaptation is necessary.

Another interesting aspect to consider when using LLM embeddings is how to best represent the class labels. A simple approach is to use one-hot encoded vectors for each class, but this does not leverage the semantic information contained in the label text. Instead, we can use the LLM to generate embeddings for each label, which can then be used in conjunction with the instance embeddings. As discussed in~\Cref{sec:kernel_regression},~\method{} uses the embedding of the label to compute a weighted sum of the context labels to make a final prediction. However, it is also possible to use $w_i$ directly as weights for the one-hot encoded label of each context -- i.e. a weighted voting classifier. To determine whether using label embeddings provides an advantage, we compare the performance of the two versions as in~\Cref{tab:oracle-label-comparison}. We find that using label embeddings significantly improves the performance of the oracle classifier over all $k$ values, showing that using the label embeddings instead of one-hot encodings provides a meaningful advantage.

\subsection{Impact of LLM choice}

\begin{table*}
	\centering
	\caption{Comparison of average balanced accuracy score across different language models.}
	\label{tab:lm-backbone-comparison}
	{
		\small
		\begin{tabular}{lrrrrrrrrr}
			\toprule
			\multirow{2}{*}{Model} & \multicolumn{9}{c}{$k$ (number of shots)}                                                                                                                                                                 \\
			                       & 2                                         & 4                 & 8                 & 16                & 32                & 64                & 128               & 256               & 512               \\
			\midrule
			Granite3.3-Base        & 74.27                                     & \underline{77.36} & 78.43             & 77.28             & 78.57             & 77.85             & 80.42             & 79.51             & 80.52             \\
			Granite3.3-Inst.       & 73.16                                     & 74.94             & 79.56             & 80.53             & 80.54             & \underline{83.52} & 81.84             & 84.27             & \underline{84.36} \\
			Llama3.1-8B            & \underline{77.18}                         & 76.90             & \underline{80.12} & \underline{82.01} & \textbf{83.76}    & \textbf{84.68}    & \textbf{86.07}    & \textbf{85.40}    & \textbf{86.32}    \\
			Qwen3                  & 57.43                                     & 56.77             & 55.42             & 56.45             & 54.92             & 56.95             & 56.81             & 56.48             & 55.66             \\
			Qwen3-Base             & 61.53                                     & 61.75             & 62.72             & 63.29             & 64.30             & 64.81             & 64.54             & 65.04             & 64.68             \\
			Qwen3-Emb.             & 73.90                                     & 75.82             & 75.69             & 78.11             & 77.01             & 78.14             & 77.72             & 78.43             & 79.10             \\
			all-MiniLM-L6-v2       & 74.24                                     & 75.26             & 76.21             & 77.84             & 76.48             & 77.57             & 76.14             & 77.44             & 77.29             \\
			TabuLa-8B              & \textbf{77.19}                            & \textbf{77.42}    & \textbf{80.57}    & \textbf{82.20}    & \underline{82.43} & 82.60             & \underline{84.69} & \underline{84.81} & 83.35             \\
			\bottomrule
		\end{tabular}
	}
\end{table*}%
Having confirmed that the predictive signal is present in the raw LLM embeddings, we investigate the \textit{impact} of using different LLM backbones. We compare the performance of the oracle classifier using embeddings from different LLMs, including TabuLa-8B, Granite3.3 (Base and Instruction-tuned), Llama3.1-8B, Qwen3 (Base, Instruction-tuned, and Embedding-tuned), and a smaller encoder-only model from Sentence Transformers (all-MiniLM-L6-v2). The results can be seen in in~\Cref{tab:lm-backbone-comparison}. We find that while all models perform reasonably well, there is a significant variation in performance across different models. Interestingly, we observe that MiniLM, which has only $\sim 22$M parameters, shows a stronger performance compared to the much larger Qwen models, especially notable in lower $k$ values. Another observation is that embedding-tuning significantly improves the performance of the Qwen model variants, although Llama and TabuLa show the strongest performance overall. In particular, we find that Llama3.1-8B shows generally the strongest performance, but it's table-fine-tuned variant, TabuLa-8B, also shows very strong performance, especially in the lower $k$ regimes.

\subsection{The Power of $\gamma$ Calibration}%
\begin{figure}
	\centering
	\includegraphics[width=\linewidth]{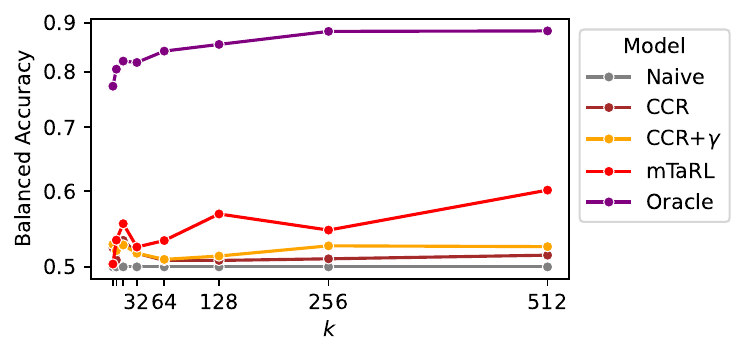}
	\captionof{figure}{
		Balanced accuracy of classifiers built from LLM embeddings as more components are added.
	}
	\label{fig:ablation-components}
\end{figure}%
We then construct the most simple classifier as described in~\Cref{sec:kernel_regression}, which we refer to as the \textit{naive} baseline. As shown in~\Cref{fig:ablation-components}, this naive approach performs poorly, barely above random chance. This confirms that a straightforward application of LLM embeddings is insufficient for tabular classification. However, it also raises the question of whether the poor performance is due to a fundamental lack of signal in the embeddings, or whether it is due to a failure to properly adapt the classifier to the task at hand. The results of~\Cref{fig:knn_vs_embedding} suggest that the embeddings do contain useful information which can be extracted to some degree by a KNN classifier, suggesting that the \textit{naive} kernel classifier alone is not sufficient to unlock the information contained in the embeddings. The results of applying the CCR and adjustment $\gamma$ confirm this hypothesis. As can be seen in~\Cref{fig:ablation-components}, applying CCR and $\gamma$ adjustments shows a boost in performance. However, even with both components, the performance is still far from satisfactory, indicating that further adaptation is necessary. Our prototype meta learning approach shows that it is indeed possible to further close the gap to the oracle classifier, again demonstrating that task-specific adaptation is crucial for good performance.

\subsection{Semantic vs. Non-Semantic Datasets}
\begin{figure}
	\centering
	\begin{subfigure}[b]{0.48\textwidth}
		\centering
		\includegraphics[width=\linewidth]{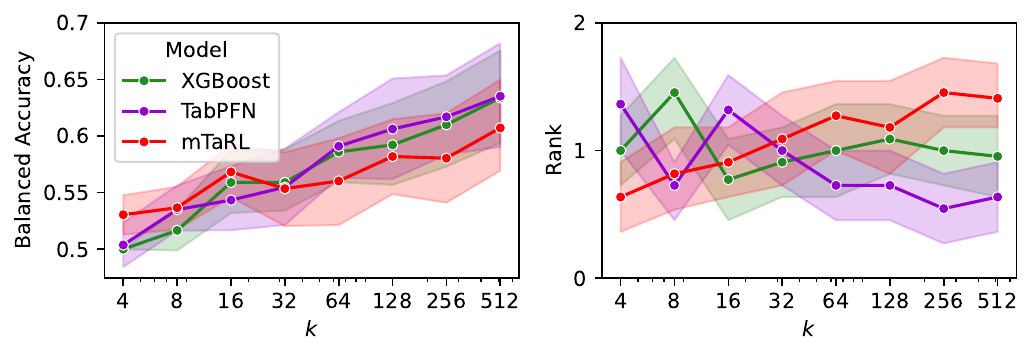}
		\caption{CARTE-Binary benchmark.}
		\label{fig:meta_vs_baseline_carte}
	\end{subfigure}
	\hfill
	\begin{subfigure}[b]{0.48\textwidth}
		\centering
		\includegraphics[width=\linewidth]{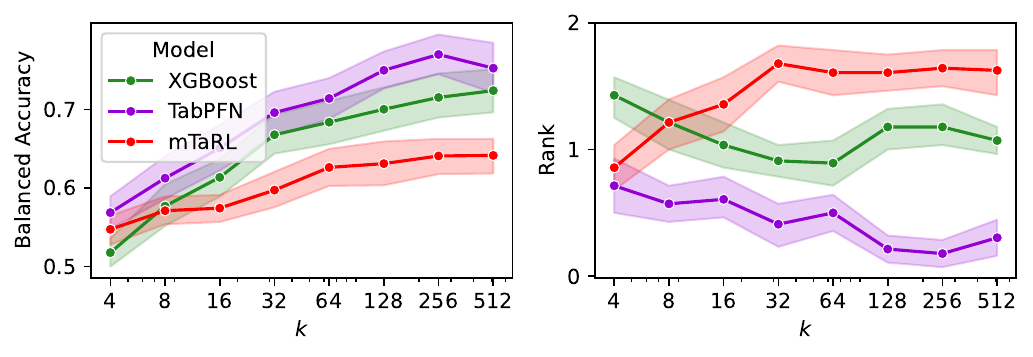}
		\caption{TabArena-Binary benchmark.}
		\label{fig:meta_vs_baseline_tabarena}
	\end{subfigure}
	\label{fig:semantic-vs-nonsemantic}
	\caption{Comparison of \metamethod{} against non-semantic baselines on (a) CARTE-CLF and (b) TabArena-CLF benchmarks. Higher is better for balanced accuracy (left) and lower is better for rank (right).}
\end{figure}%
An advantage of using LLM-based methods on tabular data is in the ability to leverage the semantic content of the tables, such as column names and values. To better understand the impact of semantic content in the downstream performance, we separately analyze the performance of non-semantic baselines (XGboost, TabPFN) against \metamethod{} on two groups of datasets: CARTE-CLF, which contains rich semantic information, and TabArena-CLF, which contains mostly non-semantic data. As shown in~\Cref{fig:semantic-vs-nonsemantic}, we find that \metamethod outperforms the non-semantic baselines on CARTE-CLF in $k \leq 32$, while non-semantic baselines show strong performance on TabArena-CLF. This suggests that the semantic content of the tables does indeed play a significant role in the effectiveness of LLM-based methods for tabular classification.

\subsection{Runtime}

\begin{figure}
	\centering
	\includegraphics[width=\linewidth]{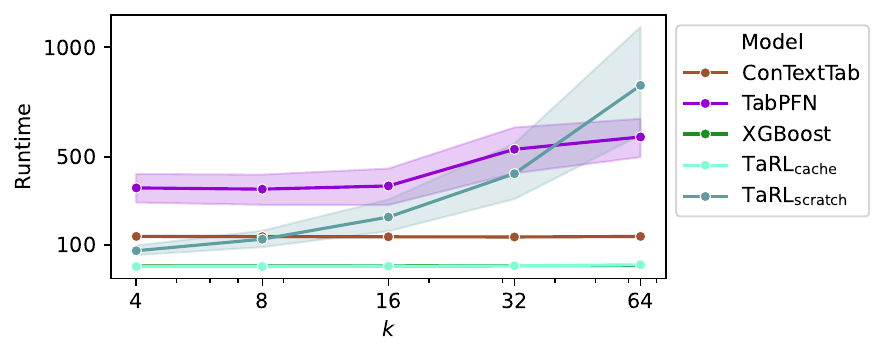}
	\caption{
		Comparison of runtime (in seconds) for different models across various $k$ values on CARTE-CLF benchmark. TabuLa-8B is excluded because its runtimes are significantly higher (up to 1,000 times) than the other models.
	}
	\label{fig:runtime-comparison}
\end{figure}%
An important consideration when deploying these models in real-life is the runtime performance. We first show in~\Cref{tab:main_runtime_comparison} that our proposed method reduces the runtime by as much as 1,000 times while showing stronger performance than using TabuLa in an auto-regressive manner.~\Cref{fig:runtime-comparison} further breaks down the runtime performance of different models across various $k$ values on CARTE-CLF benchmark. For a fair comparison, we evaluate two variants of~\method{}: Embed every sample from scratch (\method{}$_\text{scratch}$), or using a caching scheme to cache all examples ahead of time (\method{}$_\text{cache}$). We see that for up $k \leq 8$, even the \method{}$_\text{scratch}$ variant is quite fast, placing above XGBoost. However, as context size grows, the runtime of \method{}$_\text{scratch}$ increases significantly due to the repeated embedding computations. In contrast, \method{}$_\text{cache}$ remains extremely fast across all $k$ values, as the only computation necessary is to calculate the support set weights and performing the predictions. This suggests that in a real-world scenario where an LLM is already deployed and the embeddings can be cached, our method can provide an extremely fast and efficient solution for few-shot tabular classification.

\section{Conclusion}

In this work, we show that web-native, semantically rich table data can be classified few-shot by reusing off-the-shelf LLM embeddings of serialized rows, provided two light semantic adaptations: (1) common-component removal and (2) task-wise temperature calibration, closing much of the gap to specialized tabular FMs while being orders-of-magnitude cheaper. Our experiments demonstrate that~\metamethod{} achieves comparable performance to specialized state-of-the-art tabular models in low-data regimes, especially when the datasets contain meaningful semantic content. We also show that this approach is highly efficient in computation while achieving stronger performance when compared to using an LLM directly for in-context learning, making~\method{} a promising alternative to the expensive auto-regressive LLM inference.

While the primary goal of this work is to demonstrate the viability and efficiency of~\method{}, our findings can be extended in several directions. We focus on binary classification tasks in this work, but our approach can be extended to multiclass classification and regression tasks. However, further investigation is needed to fully understand the limitations and potential of our approach in these contexts. In addition, our results with the $\gamma$ parameter invite further analysis into what properties of a task determine its optimal calibration. Similarly, the success of a simple geometric correction via the CCR method motivates a more fine-grained analysis of how different tabular data structures are represented in the embedding space and whether there exist other methods that could further uncover the predictive signal in LLM embeddings.

\section*{Acknowledgments}
This work was supported by IBM through the IBM-Rensselaer Future of Computing Research Collaboration.

\balance
\bibliographystyle{ACM-Reference-Format}
\bibliography{references}
\appendix
\section{Appendix}
\label{sec:apdx}

\subsection{Reproducing Results in Paper}
\label{sec:apdx:reproducibility}
An anonymized version of our source code can be found in \url{https://anonymous.4open.science/r/tarlm-supplementary-F125/}. We will release the full code upon publication.

\subsubsection{Implementation Details}
\label{sec:apdx:implementation_details}

All experiments were conducted on a single NVIDIA H100 GPU. We use the Hugging Face Transformers library~\citep{wolfHuggingFacesTransformersStateoftheart2020} for all LLM-related computations. For the baseline models, we use the implementations from scikit-learn~\citep{pedregosaScikitlearnMachineLearning2011} and XGBoost~\citep{chenXGBoostScalableTree2016}.

\subsection{Dataset Statistics}
\label{sec:apdx:dataset_statistics}

\begin{table*}
	\centering
	\caption{
		Statistics of datasets used in the experiments.
	}
	\label{tab:dataset_statistics}
	{\tiny
		\begin{tabular}{llrrrrr}
			\toprule
			   & Dataset                                             & Benchmark           & \# classes & \# rows & \# cat. features & \# cont. features \\
			\midrule
			0  & \texttt{chocolate-bar-ratings}                      & CARTE Binary        & 2          & 2588    & 6                & 2                 \\
			1  & \texttt{coffee-ratings}                             & CARTE Binary        & 2          & 2077    & 8                & 0                 \\
			2  & \texttt{michelin}                                   & CARTE Binary        & 2          & 6832    & 5                & 2                 \\
			3  & \texttt{nba-draft}                                  & CARTE Binary        & 2          & 1669    & 4                & 2                 \\
			4  & \texttt{ramen-ratings}                              & CARTE Binary        & 2          & 4084    & 4                & 0                 \\
			5  & \texttt{roger-ebert}                                & CARTE Binary        & 2          & 2687    & 6                & 1                 \\
			6  & \texttt{spotify}                                    & CARTE Binary        & 2          & 41099   & 6                & 11                \\
			7  & \texttt{us-accidents-severity}                      & CARTE Binary        & 2          & 21656   & 7                & 1                 \\
			8  & \texttt{whisky}                                     & CARTE Binary        & 2          & 1812    & 6                & 0                 \\
			9  & \texttt{yelp}                                       & CARTE Binary        & 2          & 64616   & 8                & 3                 \\
			10 & \texttt{zomato}                                     & CARTE Binary        & 2          & 61441   & 7                & 1                 \\
			11 & \texttt{blood-transfusion-service-center}           & TabArena Binary     & 2          & 748     & 0                & 4                 \\
			12 & \texttt{Amazon-employee-access}                     & TabArena Binary     & 2          & 32769   & 9                & 0                 \\
			13 & \texttt{APSFailure}                                 & TabArena Binary     & 2          & 76000   & 1                & 168               \\
			14 & \texttt{bank-marketing}                             & TabArena Binary     & 2          & 45211   & 5                & 5                 \\
			15 & \texttt{Bank-Customer-Churn}                        & TabArena Binary     & 2          & 10000   & 3                & 4                 \\
			16 & \texttt{anneal}                                     & TabArena Multiclass & 5          & 898     & 19               & 3                 \\
			17 & \texttt{credit-card-clients-default}                & TabArena Binary     & 2          & 30000   & 8                & 14                \\
			18 & \texttt{customer-satisfaction-in-airline}           & TabArena Binary     & 2          & 129880  & 15               & 4                 \\
			19 & \texttt{coil2000-insurance-policies}                & TabArena Binary     & 2          & 9822    & 80               & 0                 \\
			20 & \texttt{diabetes}                                   & TabArena Binary     & 2          & 768     & 1                & 7                 \\
			21 & \texttt{credit-g}                                   & TabArena Binary     & 2          & 1000    & 14               & 3                 \\
			22 & \texttt{Diabetes130US}                              & TabArena Binary     & 2          & 71518   & 36               & 3                 \\
			23 & \texttt{churn}                                      & TabArena Binary     & 2          & 5000    & 3                & 14                \\
			24 & \texttt{GiveMeSomeCredit}                           & TabArena Binary     & 2          & 150000  & 4                & 6                 \\
			25 & \texttt{hazelnut-spread-contaminant-detection}      & TabArena Binary     & 2          & 2400    & 0                & 30                \\
			26 & \texttt{E-CommereShippingData}                      & TabArena Binary     & 2          & 10999   & 6                & 3                 \\
			27 & \texttt{heloc}                                      & TabArena Binary     & 2          & 10459   & 6                & 17                \\
			28 & \texttt{Fitness-Club}                               & TabArena Binary     & 2          & 1500    & 3                & 2                 \\
			29 & \texttt{HR-Analytics-Job-Change-of-Data-Scientists} & TabArena Binary     & 2          & 19158   & 9                & 2                 \\
			30 & \texttt{in-vehicle-coupon-recommendation}           & TabArena Binary     & 2          & 12684   & 17               & 0                 \\
			31 & \texttt{Is-this-a-good-customer}                    & TabArena Binary     & 2          & 1723    & 6                & 4                 \\
			32 & \texttt{maternal-health-risk}                       & TabArena Multiclass & 3          & 1014    & 4                & 2                 \\
			33 & \texttt{online-shoppers-intention}                  & TabArena Binary     & 2          & 12330   & 8                & 8                 \\
			34 & \texttt{Marketing-Campaign}                         & TabArena Binary     & 2          & 2240    & 9                & 10                \\
			35 & \texttt{kddcup09-appetency}                         & TabArena Binary     & 2          & 50000   & 77               & 123               \\
			36 & \texttt{NATICUSdroid}                               & TabArena Binary     & 2          & 7491    & 0                & 0                 \\
			37 & \texttt{students-dropout-and-academic-success}      & TabArena Multiclass & 3          & 4424    & 16               & 12                \\
			38 & \texttt{polish-companies-bankruptcy}                & TabArena Binary     & 2          & 5910    & 0                & 64                \\
			39 & \texttt{seismic-bumps}                              & TabArena Binary     & 2          & 2584    & 6                & 6                 \\
			40 & \texttt{qsar-biodeg}                                & TabArena Binary     & 2          & 1054    & 19               & 19                \\
			41 & \texttt{SDSS17}                                     & TabArena Multiclass & 3          & 78053   & 3                & 8                 \\
			42 & \texttt{splice}                                     & TabArena Multiclass & 3          & 3190    & 60               & 0                 \\
			43 & \texttt{website-phishing}                           & TabArena Multiclass & 3          & 1353    & 7                & 0                 \\
			44 & \texttt{MIC}                                        & TabArena Multiclass & 8          & 1699    & 93               & 11                \\
			45 & \texttt{jm1}                                        & TabArena Binary     & 2          & 10885   & 0                & 21                \\
			46 & \texttt{anime-planet}                               & CR                  & 1          & 14391   & 5                & 4                 \\
			47 & \texttt{babies-r-us}                                & CR                  & 1          & 5085    & 3                & 0                 \\
			48 & \texttt{beer-ratings}                               & CR                  & 1          & 3197    & 5                & 14                \\
			49 & \texttt{bikedekho}                                  & CR                  & 1          & 4786    & 5                & 2                 \\
			50 & \texttt{bikewale}                                   & CR                  & 1          & 8992    & 5                & 2                 \\
			51 & \texttt{buy-buy-baby}                               & CR                  & 1          & 10718   & 3                & 0                 \\
			52 & \texttt{cardekho}                                   & CR                  & 1          & 37813   & 12               & 2                 \\
			53 & \texttt{clear-corpus}                               & CR                  & 1          & 4724    & 12               & 13                \\
			54 & \texttt{company-employees}                          & CR                  & 1          & 10941   & 7                & 0                 \\
			55 & \texttt{employee-remuneration}                      & CR                  & 1          & 35396   & 2                & 2                 \\
			56 & \texttt{employee-salaries}                          & CR                  & 1          & 9211    & 4                & 2                 \\
			57 & \texttt{fifa22-players}                             & CR                  & 1          & 18085   & 9                & 8                 \\
			58 & \texttt{filmtv-movies}                              & CR                  & 1          & 41205   & 6                & 3                 \\
			59 & \texttt{journal-jcr}                                & CR                  & 1          & 9615    & 4                & 5                 \\
			60 & \texttt{journal-sjr}                                & CR                  & 1          & 27931   & 9                & 0                 \\
			61 & \texttt{jp-anime}                                   & CR                  & 1          & 15535   & 10               & 5                 \\
			62 & \texttt{k-drama}                                    & CR                  & 1          & 1239    & 8                & 4                 \\
			63 & \texttt{mlds-salaries}                              & CR                  & 1          & 10456   & 7                & 1                 \\
			64 & \texttt{movies}                                     & CR                  & 1          & 7224    & 7                & 6                 \\
			65 & \texttt{museums}                                    & CR                  & 1          & 11467   & 14               & 2                 \\
			66 & \texttt{mydramalist}                                & CR                  & 1          & 3400    & 10               & 3                 \\
			67 & \texttt{prescription-drugs}                         & CR                  & 1          & 1714    & 5                & 2                 \\
			68 & \texttt{rotten-tomatoes}                            & CR                  & 1          & 7158    & 9                & 4                 \\
			69 & \texttt{us-accidents-counts}                        & CR                  & 1          & 22623   & 6                & 0                 \\
			70 & \texttt{us-presidential}                            & CR                  & 1          & 19857   & 6                & 0                 \\
			71 & \texttt{used-cars-24}                               & CR                  & 1          & 5918    & 5                & 2                 \\
			72 & \texttt{used-cars-benz-italy}                       & CR                  & 1          & 16391   & 2                & 3                 \\
			73 & \texttt{used-cars-dot-com}                          & CR                  & 1          & 4009    & 7                & 2                 \\
			74 & \texttt{used-cars-pakistan}                         & CR                  & 1          & 72655   & 2                & 3                 \\
			75 & \texttt{used-cars-saudi-arabia}                     & CR                  & 1          & 5507    & 6                & 3                 \\
			76 & \texttt{videogame-sales}                            & CR                  & 1          & 16410   & 4                & 1                 \\
			77 & \texttt{wikiliq-beer}                               & CR                  & 1          & 13461   & 7                & 2                 \\
			78 & \texttt{wikiliq-spirit}                             & CR                  & 1          & 12275   & 5                & 2                 \\
			79 & \texttt{wina-pl}                                    & CR                  & 1          & 2247    & 11               & 2                 \\
			80 & \texttt{wine-dot-com-prices}                        & CR                  & 1          & 15254   & 7                & 2                 \\
			81 & \texttt{wine-dot-com-ratings}                       & CR                  & 1          & 4095    & 7                & 2                 \\
			82 & \texttt{wine-enthusiasts-prices}                    & CR                  & 1          & 120975  & 8                & 1                 \\
			83 & \texttt{wine-enthusiasts-ratings}                   & CR                  & 1          & 129971  & 8                & 1                 \\
			84 & \texttt{wine-vivino-price}                          & CR                  & 1          & 13834   & 5                & 2                 \\
			85 & \texttt{wine-vivino-rating}                         & CR                  & 1          & 13834   & 6                & 2                 \\
			86 & \texttt{airfoil-self-noise}                         & TR                  & 1          & 1503    & 3                & 2                 \\
			87 & \texttt{Another-Dataset-on-used-Fiat-500}           & TR                  & 1          & 1538    & 3                & 4                 \\
			88 & \texttt{concrete-compressive-strength}              & TR                  & 1          & 1030    & 1                & 7                 \\
			89 & \texttt{diamonds}                                   & TR                  & 1          & 53940   & 3                & 6                 \\
			90 & \texttt{healthcare-insurance-expenses}              & TR                  & 1          & 1338    & 2                & 2                 \\
			91 & \texttt{Food-Delivery-Time}                         & TR                  & 1          & 45451   & 3                & 6                 \\
			92 & \texttt{houses}                                     & TR                  & 1          & 20640   & 0                & 8                 \\
			93 & \texttt{miami-housing}                              & TR                  & 1          & 13776   & 2                & 12                \\
			94 & \texttt{physiochemical-protein}                     & TR                  & 1          & 45730   & 0                & 9                 \\
			95 & \texttt{QSAR-fish-toxicity}                         & TR                  & 1          & 907     & 2                & 4                 \\
			96 & \texttt{superconductivity}                          & TR                  & 1          & 21263   & 2                & 79                \\
			97 & \texttt{wine-quality}                               & TR                  & 1          & 6497    & 0                & 11                \\
			\bottomrule
		\end{tabular}
	}
\end{table*}

As noted in the main text, we use every dataset from the CARTE benchmark~\cite{kimCARTEPretrainingTransfer2024}, and all but 4 datasets from the TabArena benchmark~\cite{ericksonTabArenaLivingBenchmark2025}. The 4 excluded datasets were omitted due to their large size, which made them impractical for our few-shot experiments which requires repeated sampling of small subsets.

\subsection{Detailed results}
\label{sec:apdx:detailed_results}

Here we provide the results for regression tasks in addition to the classification results presented in the main text.

\begin{table*}
	\centering
	\caption{Detailed comparison of average balanced accuracy score per benchmark. Best performance in each benchmark/$k$ combination is marked with \textbf{bold} and the second best with \underline{underline}.}
	\label{tab:detailed-comparison-reg}
	{\tiny
		\begin{tabular}{ll p{0.8cm}p{0.8cm} *{8}{r}}
			\toprule
			\multirow{2}{*}{Benchmark} & \multirow{2}{*}{Model} & \multicolumn{2}{c}{Rank} & \multicolumn{8}{c}{$k$ (number of shots)}                                                                                                                                                                         \\
			                           &                        & $k \leq 32$              & $k > 32$                                  & 4                  & 8                  & 16                 & 32                 & 64                 & 128                & 256                & 512                \\
			\cmidrule(lr){1-2} \cmidrule(lr){3-4} \cmidrule(lr){5-12}

			\multirow{7}{*}{\shortstack{CARTE                                                                                                                                                                                                                                                                  \\
			Regression (40)}}          & KNN                    & 6.8125                   & 7.9562                                    & 0.2803             & 0.2814             & 0.2985             & 0.2985             & 0.3139             & 0.3223             & 0.3271             & 0.3355             \\
			                           & KNN-Emb.               & 7.2750                   & 7.5000                                    & 0.2539             & 0.2756             & 0.2918             & 0.3076             & 0.3218             & 0.3469             & 0.3628             & 0.3766             \\
			                           & XGBoost                & 6.1625                   & 5.7625                                    & 0.2437             & 0.3004             & 0.3441             & 0.3763             & 0.3997             & 0.4386             & 0.4637             & 0.4843             \\
			                           & TabPFN                 & \textbf{4.3688}          & \textbf{3.7687}                           & \textbf{0.3147}    & \textbf{0.3237}    & \textbf{0.3740}    & \textbf{0.4174}    & \textbf{0.4394}    & \textbf{0.4831}    & \textbf{0.5118}    & \textbf{0.5472}    \\
			                           & ConTextTab             & \underline{5.1125}       & \underline{4.1125}                        & \underline{0.2993} & \underline{0.3195} & \underline{0.3537} & \underline{0.3956} & \underline{0.4278} & \underline{0.4810} & \underline{0.5100} & \underline{0.5311} \\
			                           & \metamethod{}          & 6.5563                   & 7.4750                                    & 0.2767             & 0.2965             & 0.3078             & 0.3158             & 0.3151             & 0.3509             & 0.3710             & 0.3737             \\
			                           & \gray{Oracle}          & \gray{0.5844}            & \gray{1.1875}                             & \gray{0.5021}      & \gray{0.5230}      & \gray{0.5393}      & \gray{0.5554}      & \gray{0.5759}      & \gray{0.5925}      & \gray{0.6097}      & \gray{0.6210}      \\ \midrule

			\multirow{7}{*}{\shortstack{TabArena                                                                                                                                                                                                                                                               \\
			Multiclass (6)}}           & KNN                    & 7.0714                   & 7.2143                                    & 0.2940             & 0.3388             & 0.3823             & 0.4031             & 0.4352             & 0.5199             & 0.5313             & 0.4930             \\
			                           & KNN-Emb.               & 7.7857                   & 7.9286                                    & 0.2876             & 0.3594             & 0.4029             & 0.4363             & 0.4574             & 0.4972             & 0.5623             & 0.5511             \\
			                           & XGBoost                & 6.3214                   & 3.9643                                    & 0.2190             & 0.4352             & 0.5262             & 0.6323             & 0.6667             & 0.7019             & 0.7516             & \underline{0.7762} \\
			                           & TabPFN                 & \underline{4.3929}       & \textbf{2.0893}                           & \textbf{0.3993}    & \underline{0.4677} & \underline{0.5508} & \underline{0.6521} & \underline{0.7023} & \underline{0.7567} & \textbf{0.7945}    & \textbf{0.8109}    \\
			                           & ConTextTab             & \textbf{3.9107}          & \underline{2.9821}                        & \underline{0.3561} & \textbf{0.5242}    & \textbf{0.5562}    & \textbf{0.6709}    & \textbf{0.7050}    & \textbf{0.7686}    & \underline{0.7697} & 0.7692             \\
			                           & \metamethod{}          & 6.8750                   & 7.6429                                    & 0.3263             & 0.4260             & 0.4218             & 0.4729             & 0.4961             & 0.5354             & 0.5863             & 0.5867             \\
			                           & \gray{Oracle}          & \gray{1.6964}            & \gray{3.6429}                             & \gray{0.4402}      & \gray{0.6150}      & \gray{0.6366}      & \gray{0.6690}      & \gray{0.6781}      & \gray{0.6969}      & \gray{0.7455}      & \gray{0.7403}      \\ \midrule

			\multirow{7}{*}{\shortstack{TabArena                                                                                                                                                                                                                                                               \\
			Regression (13)}}          & KNN                    & 6.0625                   & 7.6042                                    & 0.2917             & 0.3410             & 0.3736             & 0.3858             & 0.3953             & 0.4404             & 0.4689             & 0.4745             \\
			                           & KNN-Emb.               & 7.6562                   & 8.1875                                    & 0.2638             & 0.2829             & 0.2918             & 0.3238             & 0.3674             & 0.3911             & 0.4320             & 0.4671             \\
			                           & XGBoost                & 6.1667                   & 5.0000                                    & 0.2292             & 0.3498             & 0.4318             & 0.4613             & 0.5110             & 0.5703             & 0.5978             & 0.6381             \\
			                           & TabPFN                 & \textbf{3.2917}          & \textbf{2.6667}                           & \textbf{0.3472}    & \textbf{0.4042}    & \textbf{0.4830}    & \textbf{0.5340}    & \underline{0.5843} & \textbf{0.6005}    & \textbf{0.6572}    & \textbf{0.7011}    \\
			                           & ConTextTab             & \underline{4.0000}       & \underline{3.2917}                        & \underline{0.3348} & \underline{0.3906} & \underline{0.4600} & \underline{0.5200} & \textbf{0.5901}    & \underline{0.5845} & \underline{0.6341} & \underline{0.6624} \\
			                           & \metamethod{}          & 7.0417                   & 7.8958                                    & 0.2715             & 0.3079             & 0.3509             & 0.3533             & 0.3987             & 0.4291             & 0.4457             & 0.4692             \\
			                           & \gray{Oracle}          & \gray{0.9583}            & \gray{1.8125}                             & \gray{0.4836}      & \gray{0.5143}      & \gray{0.5570}      & \gray{0.5768}      & \gray{0.6289}      & \gray{0.6525}      & \gray{0.6619}      & \gray{0.6955}      \\

			\bottomrule
		\end{tabular}
	}
\end{table*}

\subsection{Meta-Learner implementation details}
\label{sec:apdx:meta_learnre_details}

\begin{algorithm}[H]
	{\small
		\caption{Meta-Training Dataset Generation}
		\label{alg:meta_datagen_compact}
		\begin{algorithmic}[1]
			\State \textbf{Input:} Meta-training datasets $\mathcal{D}_{\text{meta-train}}$, support sizes $\mathcal{K}$, tasks per setting $T$, search space $\Gamma_{\text{search}}$, feature extractor $f$.
			\State \textbf{Initialize:} Meta-datasets $\{\mathcal{X}_k, \mathcal{Y}_k\}_{k \in \mathcal{K}}$ as empty.
			\For{each dataset $D \in \mathcal{D}_{\text{meta-train}}$ and each size $k \in \mathcal{K}$}
			\State Sample $T$ tasks $\{(\gS_i, (\rx_{q,i}, y_{q,i}))\}_{i=1}^T$ from $D$, where $|\gS_i| = k$.
			\For{each task $i=1, \dots, T$} \quad {\color{gray}\footnotesize\# Find optimal $\gamma^*_i$ and construct meta-training pair}
			\State $\gamma^*_i \leftarrow \arg\max_{\gamma \in \Gamma_{\text{search}}} \mathbb{I}(\textsc{\method{}-Predict}(\gS_i, \rx_{q,i}, \gamma) = y_{q,i})$
			\State $\mathbf{z}_i \leftarrow f(\gS_i)$; Append $(\mathbf{z}_i, \gamma^*_i)$ to $(\mathcal{X}_k, \mathcal{Y}_k)$. \quad {\color{gray}\footnotesize\# extract features from context}
			\EndFor
			\EndFor
			\State \textbf{Return:} Meta-datasets $\{\mathcal{X}_k, \mathcal{Y}_k\}_{k \in \mathcal{K}}$
		\end{algorithmic}
	}
\end{algorithm}

\begin{algorithm}[H]
	{\small
		\caption{Meta-Inference with Learned Gamma Calibration}
		\label{alg:meta_inference_compact}
		\begin{algorithmic}[1]
			\State \textbf{Input:} Meta-datasets $\{\mathcal{X}_k, \mathcal{Y}_k\}_{k \in \mathcal{K}}$, a new task $(\gS_{\text{test}}, \rx_{q, \text{test}})$ with $|\gS_{\text{test}}| = k$, feature extractor $f$.
			\State Train a regression model $M_k$ on $(\mathcal{X}_k, \mathcal{Y}_k)$.
			\State Compute heuristic $\gamma_{\text{NN}}$ for the test task using Eq.~\ref{eq:gamma_heuristic}.
			\State $\mathbf{z}_{\text{test}} \leftarrow f(\gS_{\text{test}})$ \quad {\color{gray}\footnotesize\# Extract same features as $\gX_k$}
			\State $\bar{\gamma} \leftarrow M_k(\mathbf{z}_{\text{test}})$ \quad {\color{gray}\footnotesize\# Predict the corrective scale factor}
			\State $\hat{y}_q \leftarrow \textsc{\method{}-Predict}(\gS_{\text{test}}, \rx_{q, \text{test}}, \bar{\gamma})$
			\State \textbf{Return:} Prediction $\hat{y}_q$
		\end{algorithmic}
	}
\end{algorithm}

\begin{table*}[htbp]
	\centering
	\caption{Meta-features for Gamma Prediction in Few-shot Learning}
	\label{tab:meta_features}
	\resizebox{\textwidth}{!}{
		\tiny		\begin{tabular}{p{2cm}p{8cm}}
			\toprule
			\textbf{Feature Name}           & \textbf{Description}                                                                                        \\
			\midrule

			\multicolumn{2}{l}{\textbf{Basic Context Statistics}}                                                                                         \\
			\texttt{num\_classes}           & Number of unique classes.                                                                                   \\
			\texttt{minority\_class\_ratio} & Fraction of minority class.                                                                                 \\
			\texttt{majority\_class\_ratio} & Fraction of majority class.                                                                                 \\

			\midrule
			\multicolumn{2}{l}{\textbf{Context-Query Similarity Distribution Features}}                                                                   \\
			\texttt{sim}                    & Statistics of cosine similarity (normalized dot prod.) between query and context examples.                  \\
			\texttt{l2}                     & Statistics of euclidean distance between query and context examples.                                        \\

			\midrule
			\multicolumn{2}{l}{\textbf{Context-Query Similarity Distribution Features}}                                                                   \\
			\texttt{sim-mean}               & Statistics of cosine similarity (normalized dot prod.) between context examples and mean context embedding. \\
			\texttt{l2-mean}                & Statistics of euclidean distance between context examples and mean context embedding.                       \\

			\midrule
			\multicolumn{2}{l}{\textbf{$\gamma$-probing Features}}                                                                                        \\
			\texttt{majority}               & Confidence for majority label.                                                                              \\
			\texttt{minority}               & Confidence for minority label.                                                                              \\
			\texttt{entropy}                & Entropy of confidence (logits).                                                                             \\

			%

			\bottomrule
		\end{tabular}
	}
\end{table*}

\Cref{tab:meta_features} lists the meta-features used for gamma prediction. These features are computed from the context set embeddings and labels, as well as the query embedding. We use the same feature set for all values of $k$ (number of shots). The features can be categorized into three groups:
\begin{itemize}
	\item \textbf{Basic Context Statistics:} These features capture fundamental properties of the context set, such as the number of classes and class distribution.
	\item \textbf{Context-Query Similarity Distribution Features:} These features describe the distribution of similarities and distances between the query example and the context examples, as well as between context examples and the mean context embedding.
	\item \textbf{$\gamma$-probing Features:} These features are derived from the model's confidence scores (logits) when using a heuristic gamma value. They include the confidence for the majority and minority classes, as well as the entropy of the confidence distribution.
\end{itemize}
For the similarity-based features where we end up with a \textit{distribution} of values (e.g., cosine similarities between the query and each context example), we compute summary statistics by computing the following: mean, standard deviation, minimum, maximum, skewness and kurtosis.

%

\end{document}